\documentclass[letterpaper, 10 pt, conference]{ieeeconf}  

\IEEEoverridecommandlockouts                              

\usepackage{graphics} 
\usepackage{epsfig} 
\usepackage{mathptmx} 
\usepackage{times} 
\usepackage{amsmath} 
\usepackage{amssymb}  
\usepackage{siunitx}
\usepackage{subcaption}
\makeatletter
\let\NAT@parse\undefined
\makeatother
\usepackage{CJK}
\usepackage{indentfirst}
\usepackage{cases}
\usepackage{hyperref}
\usepackage{booktabs} 
\usepackage{multirow}

\title{\LARGE \bf
MAVRL: Learn to Fly in Cluttered Environments with Varying Speed
}

\author{Hang Yu$^{1}$, Christophe De Wagter$^{1}$ and Guido C. H. E de Croon$^{1}$
\thanks{All authors are with Faculty of Aerospace Engineering, Delft University of Technology, 2629 HS Delft, The Netherlands. (email: {\tt\small h.y.yu@tudelft.nl};  {\tt\small g.c.h.e.deCroon@tudelft.nl}; {\tt\small c.dewagter@tudelft.nl}).}%
}

\begin{document}

\maketitle
\thispagestyle{empty}
\pagestyle{empty}

\begin{abstract}

Many existing obstacle avoidance algorithms overlook the crucial balance between safety and agility, especially in environments of varying complexity. In our study, we introduce an obstacle avoidance pipeline based on reinforcement learning. This pipeline enables drones to adapt their flying speed according to the environmental complexity. Moreover, to improve the obstacle avoidance performance in cluttered environments, we propose a novel latent space. The latent space in this representation is explicitly trained to retain memory of previous depth map observations. Our findings confirm that varying speed leads to a superior balance of success rate and agility in cluttered environments. Additionally, our memory-augmented latent representation outperforms the latent representation commonly used in reinforcement learning. Finally, after minimal fine-tuning, we successfully deployed our network on a real drone for enhanced obstacle avoidance.

\end{abstract}


\section{INTRODUCTION}

Obstacle avoidance is a fundamental challenge in autonomous drone technology. While the past decades have seen a proliferation of obstacle avoidance algorithms \cite{Loquercio2021Science, kulkarni2023semantically, 9812231, 9345970, 10160563}, particularly those based on learning methods, their application within reinforcement learning (RL) frameworks \cite{kaufmann2023champion, singla2019memory} presents unique challenges. These methods, favored for their lower computational costs, continue to grapple with issues of performance and generalization.

In the realm of autonomous drone RL, the decision to use end-to-end learning from raw data \cite{kulkarni2023semantically, sadeghi2016cad2rl} or opt for a more efficient fixed latent space learning approach significantly impacts training efficacy and policy performance \cite{10.1007/978-3-031-47966-3_20, 9385894}. End-to-end RL provides a comprehensive learning method, yet it requires substantial computational resources and extensive datasets. On the other hand, compressing high-dimensional image data into low-dimensional latent helps improve learning efficiency. Furthermore, most studies allow robots to have relatively fixed speeds which simplifies the training process but may not perform optimally in complex environments. In contrast, variable speed offers greater adaptability but increases the complexity of training. Additionally, a critical challenge in drone technology is bridging the 'reality gap'—the translation of simulated success into real-world effectiveness. This underscores the necessity of real-world testing, as demonstrated in our study, to validate simulated findings.

\begin{figure}[thpb]
   \centering
   \subcaptionbox{}{%
   \includegraphics[width=3.0in, trim=10 280 0 0, clip]{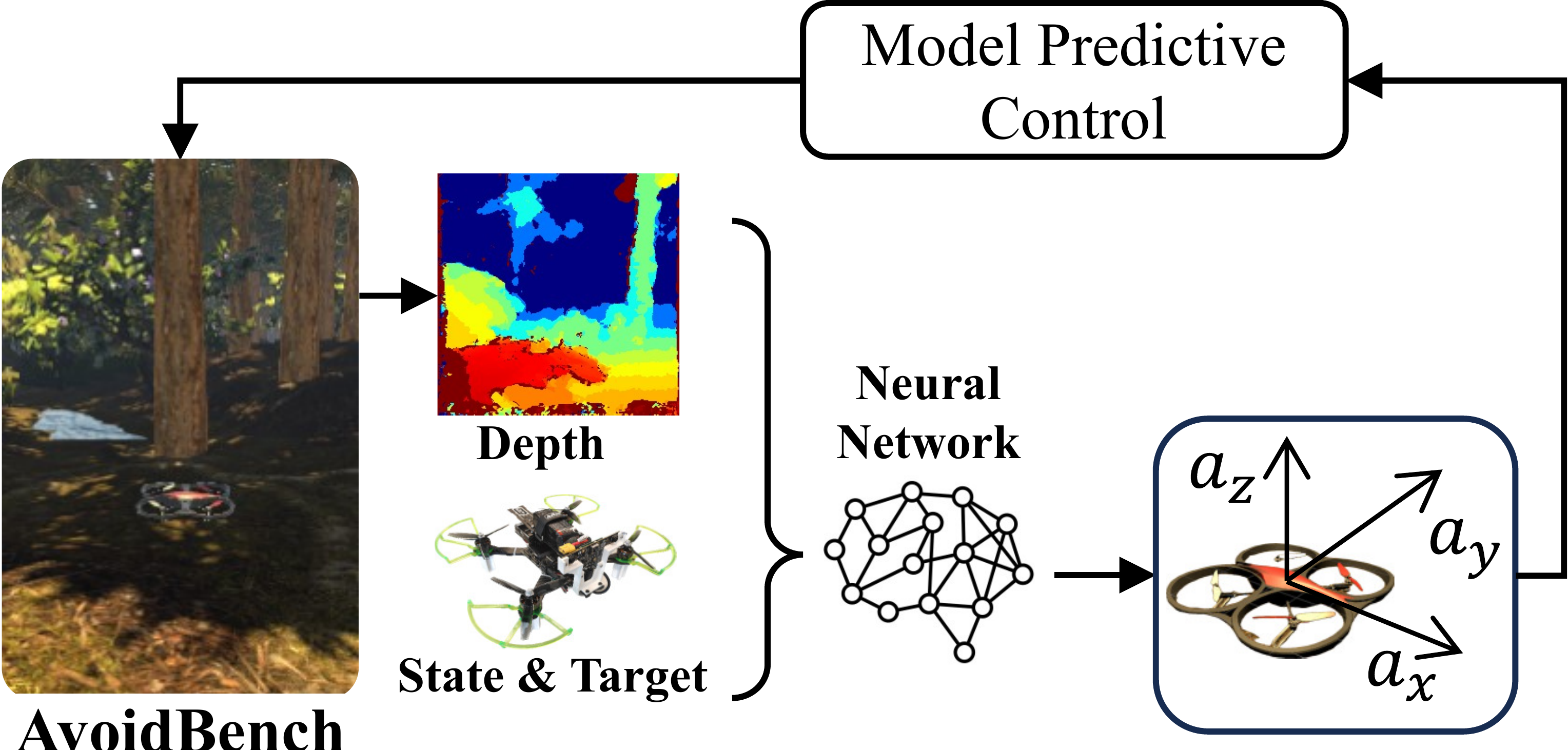}}
   \subcaptionbox{}{%
   \includegraphics[width=3.1in]{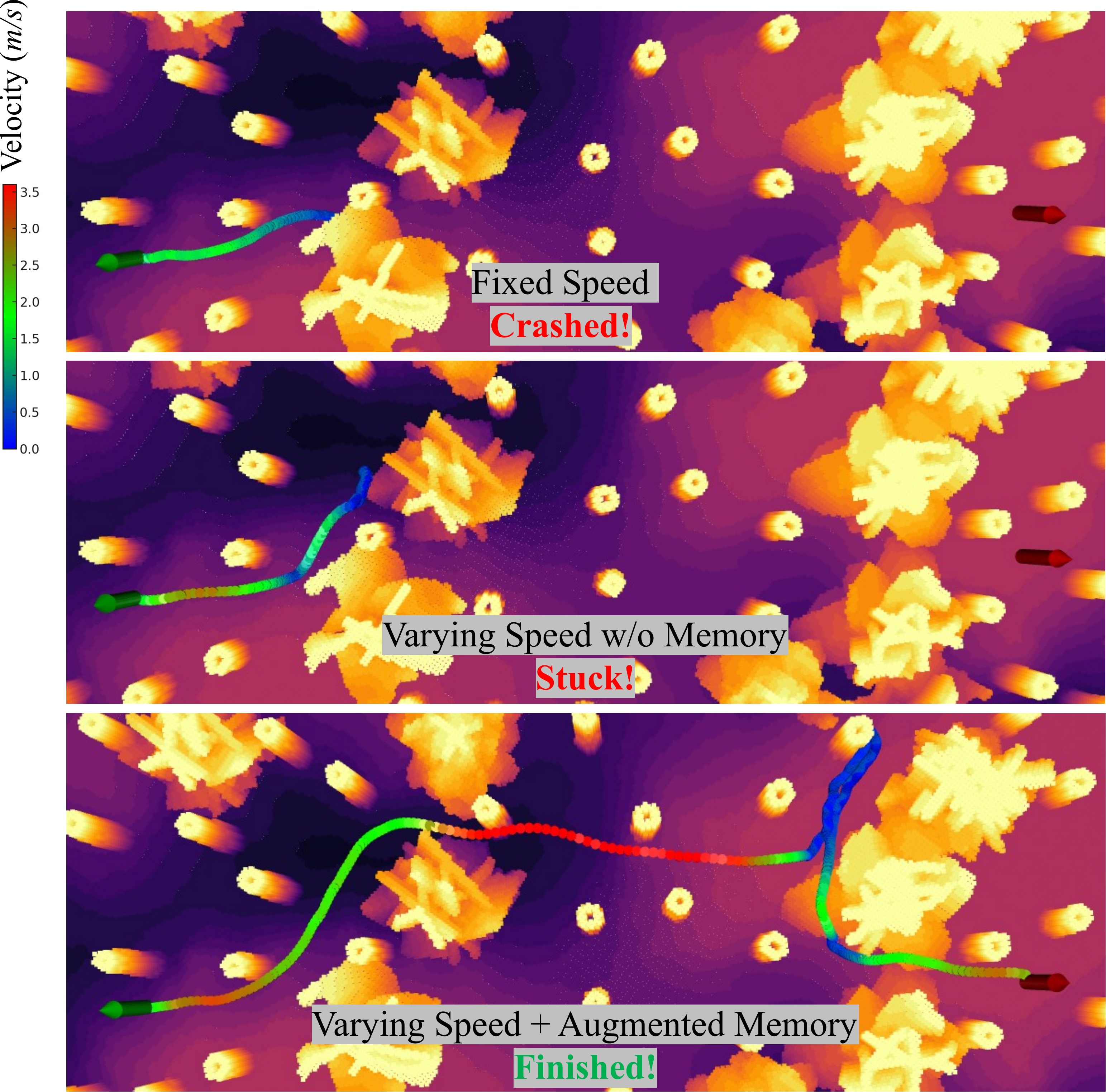}}
   \caption{(a) is the basic framework of MAVRL. (b) illustrates drone's trajectories in a Cluttered Environment. Fixed-speed flight often results in collisions with large obstacles. Absence of augmented memory leads to frequent entrapment in such obstacles. In contrast, MAVRL-equipped flights demonstrate safe and efficient navigation through complex terrains.
   }
   \label{figure1}
   \vspace{-5mm}
\end{figure}

In our study, we introduce a novel obstacle avoidance pipeline named Memory-Augmented Varying-speed Reinforcement Learning (MAVRL). As shown in Figure \ref{figure1}(a), MAVRL utilizes depth maps, along with the drone's current and target states as inputs, and generates acceleration commands as outputs. Following the acceleration generation, model predictive control (MPC) from Agilicious \cite{foehn2022agilicious} is employed to derive body rate and thrust commands for the drone. We train the pipeline in a simulated environment, featuring randomly generated obstacles of various complexities. Additionally, our approach introduces a novel latent space design that explicitly integrates memory. This latent space induces the drone to remember obstacles it has seen within a certain period of time, even if they are already outside of the field of view. As shown in Figure \ref{figure1}(b), our memory-augmented latent representation and varying speed policy enable the drone to fly in a safe and efficient way instead of getting stuck in front of large obstacles or colliding with them.

Our main contributions are as follows:
\begin{itemize}
   \item MAVRL is developed to enable adaptive-speed flight in cluttered environments, showing superior obstacle avoidance performance compared to existing state-of-the-art methods that employ constant speeds.
   \item A new latent space which retains past depth map observations is designed to improves the drone's ability to navigate around obstacles that are not currently visible but have been encountered previously.
   \item The network is effectively deployed on a real drone with minimal fine-tuning post-simulation training, demonstrating the practicality of our solution.
\end{itemize}

\section{RELATED WORK}

\subsection{Learning-based Obstacle Avoidance}

In recent years, learning-based methods for obstacle avoidance have gained significant traction, as evidenced by various studies \cite{Loquercio2021Science, kulkarni2023semantically, 9812231}. These methods generally fall into three categories: supervised learning, reinforcement learning (RL) and self-supervised learning. In the realm of supervised learning, Antonio \textit{et al.} \cite{Loquercio2021Science} introduced Agile-Autonomy, employing the Metropolis-Hastings (M-H) sampling method to calculate a distribution of collision-free trajectories, with a neural network subsequently learning the policy. Another approach \cite{9812231} also relies on learning policies from motion primitives, using a network to predict collision probabilities of the sampled trajectories.

Reinforcement learning studies \cite{song2023reaching, kaufmann2023champion} have demonstrated that training in high-fidelity simulators can exceed optimal control performance, with \cite{kaufmann2023champion} developing 'Swift,' a system surpassing champion-level human pilots. However, such methods often rely on prior knowledge or additional detection modules, restricting their effectiveness in unfamiliar environments. \cite{10160563} employed reinforcement learning to create a vision-based policy from a teacher policy with comprehensive state information. \cite{nakhleh2023sacplanner} proposed a 2D navigation planner using LiDAR-based costmaps and the Soft Actor Critic (SAC) algorithm, but the additional weight of LiDAR sensors limits practical deployment. Other methods \cite{singla2019memory, sadeghi2016cad2rl, KIM2022116742} utilize RGB images as input, transforming them into a latent format conducive to RL training. Beyond supervised and reinforcement learning, self-supervised approaches have been used for obstacle avoidance \cite{9345970, lamers2016self, gandhi2017learning, schoepe2024finding}. Notably, \cite{schoepe2024finding} introduced a robotic system that adapts its flying speed to obstacle density, drawing inspiration from flies and bees. This system employs Optic Flow (OF) to gauge environmental clutter. However, we believe we are the first to merge adaptive speed with reinforcement learning for an autonomously learned, varying-speed obstacle avoidance system.

\subsection{Latent Representations}
Given the high dimensionality of visual inputs, the role of latent representation is pivotal in effectively processing this intricate data. Studies like \cite{Loquercio2021Science, kulkarni2023semantically, sadeghi2016cad2rl} demonstrate the use of depth or RGB images to orchestrate aerial vehicle motions in an end-to-end fashion. However, such methods are not full-proof, particularly in cluttered environments, as evidenced by a sub-optimal success rate \cite{10.1007/978-3-031-47966-3_20}.
Latent representation is integral to numerous applications, including image classification \cite{SELLAMI2022108224} and vision-based navigation \cite{8653875, kulkarni2023semantically, 9385894}. Study \cite{8653875} introduced a latent representation for sampling-based motion planning. This representation incorporates AutoEncoders to encapsulate high-dimensional states like images, a dynamics network for predicting the next state, and a collision checker network. Mihir \textit{et al.} \cite{10.1007/978-3-031-47966-3_20} developed a unique collision encoding method for depth images, adept at preserving information about thin objects. When compared with a standard Variational Autoencoder (VAE) \cite{doersch2016tutorial}, their method demonstrated an ability to retain more details with the same latent dimensions. Further, \cite{9385894} unveiled a learning-based pipeline for local navigation with quadrupedal robots in cluttered settings, featuring a pre-trained state representation. This representation combines a VAE to process depth images and a Long Short-Term Memory (LSTM) network \cite{hochreiter1997long} to predict the next latent state. 

Our work draws inspiration from \cite{9385894}, but we specifically focus on enhancing the latent representation to embody more explicit past memories, rather than only predicting future states. We have validated that our approach yields superior performance, particularly in cluttered environments with large obstacles.

\section{Learning a Memory-augmented Representation}


\begin{figure*}[thpb]
   \centering
   \subcaptionbox{}{%
   \vspace{-2mm}\includegraphics[width=4.5in,trim={0 0 0 0},clip]{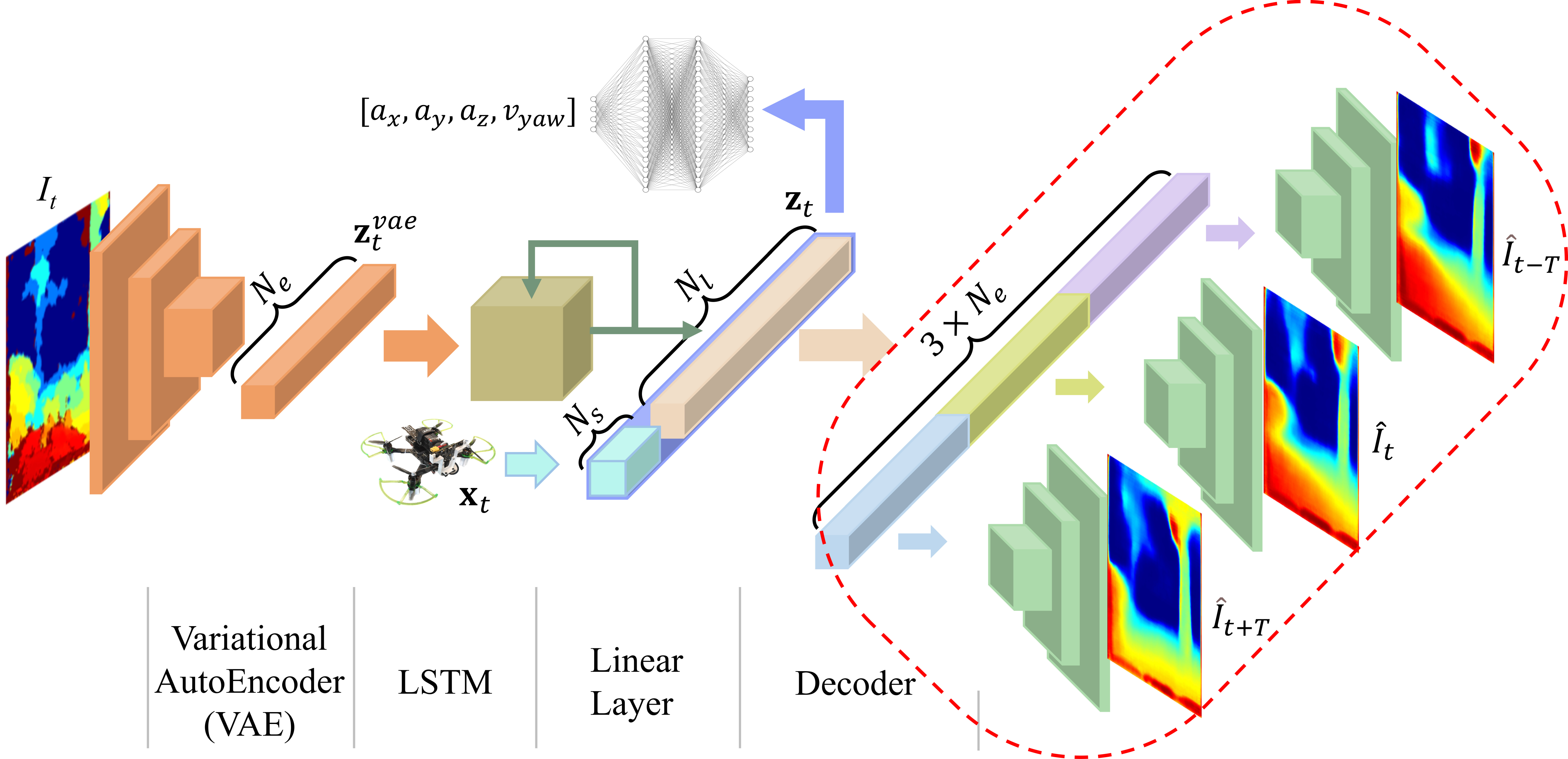}}
   \subcaptionbox{}{%
   \vspace{1mm}\includegraphics[width=2.4in,trim={0 0 0 0},clip]{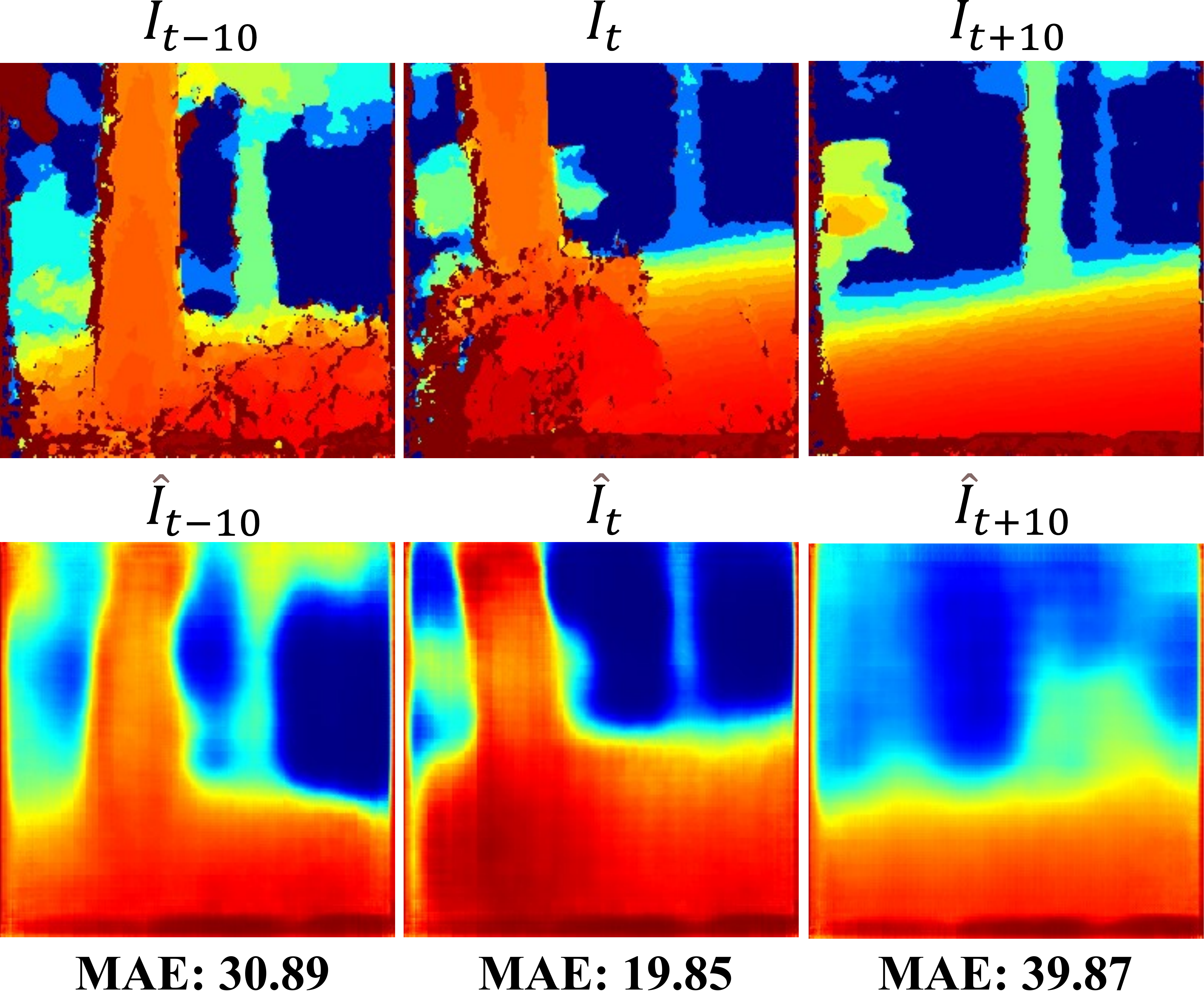}}
   \caption{(a) depicts MAVRL's network architecture. The depth image, encoded into a latent space by VAE, is processed by LSTM to create a memory-augmented representation. This, combined with the drone's state and target data, informs the acceleration command via PPO. The part of the network within the red dotted box, used for LSTM training, is not for drone deployment. (b) compares original and reconstructed depth images from latent space $\mathbf{z}_t$, showing better quality for past and current images than for future ones, due to the difficulty in predicting future states (highest MAE loss).}
   \label{figure2}
\end{figure*}

In this section, we outline our approach to learning a latent space using a 256-dimensional vector to represent depth. 
Our method focuses on encoding a sequence of depth images into this latent space, enabling the drone to retain memory of obstacles encountered over a specific time frame.

As illustrated in Figure \ref{figure2}(a), our process begins with using a VAE to convert the current depth image into an initial latent representation $\mathbf{z}_t^{vae}$. Subsequently, a LSTM network processes this sequence, generating a final latent state capable of representing past, present, or future depth images. The LSTM output, $\mathbf{z}_t$, is then merged with a vector $\mathbf{x}_t$ that encapsulates the drone's current state and target information. This combined vector is input into the Proximal Policy Optimization (PPO) \cite{schulman2017proximal} reinforcement learning algorithm, which computes the acceleration command in our case. Our pipeline comprises three main components: the VAE, LSTM, and PPO. The training process is as following:
\begin{itemize}
\item We begin by training a basic PPO policy, while the VAE and LSTM components are initially set to random. This foundational policy allows the drone to navigate 
to the target in environments without obstacles.
\item This initial policy is utilized to gather a dataset, focused primarily on capturing a multitude of depth image sequences without the concern of collisions. Subsequently, we use this dataset for the training of the VAE, bypassing the LSTM phase in this step.
\item Once the VAE is trained, we maintain the encoder in a fixed state and proceed to train the LSTM using the dataset generated by the initial policy.
\item After training both the VAE and LSTM, we freeze them and retrain the PPO, adapting it to environments of varying complexity.
\end{itemize}

In the subsequent subsections, we will delve into the specifics of training both the VAE and LSTM.

\subsection{Encoding Depth Images} 
Our pipeline is built upon AvoidBench \cite{10161097}, a high-fidelity simulator offering photo-realistic scenes. Although direct acquisition of depth images in simulations is possible, AvoidBench opts for a more realistic approach by computing these images using a semi-global matching algorithm (SGM) \cite{hirschmuller2007stereo} from a virtual stereo camera setup. This method is chosen specifically because it replicates depth errors similar to those encountered in real-world scenarios, thereby reducing the domain gap between simulation and reality. We use AvoidBench to generate a dataset of depth images, which we subsequently employ to train the VAE.

Consider a depth image at time $t$ from the dataset, denoted as $I_t \in \mathbb{D}$, where $\mathbb{D}$ represents the set of all depth images. We employ a VAE to encode each depth image into a latent space $\mathbf{z}_t^{vae} \in \mathbb{Z}^{N_e}$ $(N_e=64)$, with $\mathbb{Z}^{N_e}$ being the set of all possible latent spaces and $N_e$ the dimension of the VAE latent space. During VAE training, we focus on an encoder-decoder structure, excluding any recurrent architecture. Both the encoder and decoder consist of convolutional neural networks.

The encoder includes six convolutional layers, each followed by a ReLU activation function. The output from the final convolutional layer is flattened and then split into two components by two fully connected layers, representing the mean $\mu$ and variance $\sigma^2$. The latent space $\mathbf{z}_t^{vae}$ is sampled from a Gaussian distribution characterized by this mean $\mu$ and variance $\sigma^2$. The decoder, mirroring the encoder, comprises six deconvolutional layers, each also followed by a ReLU activation function. The output of the last deconvolutional layer passes through a sigmoid activation function to yield the reconstructed depth image $I_t^{\text{recon}}$. The loss function for the VAE is detailed in Equation \ref{eq1}.

\begin{equation}
\begin{aligned}
\mathcal{L}_{VAE} &= \mathcal{L}_{recon} + \beta_{norm} \mathcal{L}_{KL} \\
\mathcal{L}_{recon} &= {\rm MSE}(I_t, I_t^{recon}) \\
\mathcal{L}_{KL} &= \frac{1}{2}\sum_{i=1}^{N_e}(1 - \mu_i^2 - \sigma_i^2 + \log(\sigma_i^2)) \\
\end{aligned}
\label{eq1}
\end{equation}
where $\beta_{norm}$ is the weight of Kullback-Leibler (KL) loss, $I_t^{recon}$ is the reconstructed depth image from latent space $\mathbf{z}_t^{vae}$. The MSE loss is used to calculate the reconstruction loss $\mathcal{L}_{recon}$, while KL loss is used to calculate the KL divergence between the latent space and the Gaussian distribution.

\subsection{Memory-augmented Latent Representation}

As depicted in Figure \ref{figure2}(a), the output of the VAE, denoted as $\mathbf{z}_t^{vae}$, serves as the input to a single-layer LSTM network. During the LSTM training phase, its output $\mathbf{z}_t \in \mathbb{Z}^{N_l}$ $(N_l=256)$ first goes through a fully connected layer, resulting in a vector of dimension $3 \times N_e$. This vector is then divided into three segments, each corresponding to the reconstruction of past, current, and future depth images, represented as $\hat{I}_{t-T}$, $\hat{I}_t$, and $\hat{I}_{t+T}$, respectively. Notably, in our experiments, these three reconstructed images utilize the same decoder. The specific loss function employed for LSTM training is detailed in Equation \ref{eq2}.

\begin{equation}
   \begin{aligned}
   \mathcal{L}_{\text{LSTM}} = \sum_{i=-1,0,1}\lambda_i \cdot \text{MSE}(I_{t+iT}, \hat{I}_{t+iT}) \quad (\lambda_i \in \{0,1\}) \\
   \end{aligned}
   \label{eq2}
\end{equation}

where $\hat{I}_{t+iT}$ denotes the reconstructed depth image from the latent space $\mathbf{z}_t$. The coefficient $\lambda_i$ determines whether the past, current, or future depth image will be reconstructed during training. In the section V, we will show the impact of different $\lambda_i$ configurations.

Figure \ref{figure2}(b) shows depth images reconstructed from the latent space $\mathbf{z}_t$. The past and current images are more detailed than future ones, indicating the LSTM module's better encoding of past and present over future depth images. This is likely due to the unpredictability of future events and unseen environmental aspects.

\section{Reinforcement Learning for Obstacle Avoidance}

In this section, we introduce the reinforcement learning algorithm utilized for training our obstacle avoidance policy. Our approach employs PPO, a policy gradient method, optimizes the policy network by maximizing the expected reward. We frame the task of obstacle avoidance as a Markov Decision Process (MDP), offering a structured approach for decision-making in stochastic environments. Our method, distinct from other RL-based obstacle avoidance strategies \cite{10160563, nakhleh2023sacplanner}, is designed to train the policy across environments of varying complexity, enabling it to generate adaptive actions in response to environmental intricacies.

\subsection{Problem Formulation}

We employ the AvoidBench simulator \cite{10161097} for RL environment setup, which offers photo-realistic scenes and a dynamics model from RotorS \cite{furrer2016rotors}. Our drone, equipped with a stereo camera for depth image generation, utilizes PPO for training. To enhance training efficiency, we substitute RotorS with a simpler kinematics model, allowing parallel data generation with multiple drones in the same scene. The control command comprises 3D acceleration in the body frame and a 1D yaw rate. The kinematics model is:
\begin{equation}
   \begin{aligned}
   \dot{p} &= R_b^w v, \quad
   \dot{v} &= a,
   \end{aligned}
   \label{eq3}
\end{equation}
where $p$ is the drone's position in the world frame, $v$ is its velocity in the body frame, $R_b^w$ is the rotation matrix from the body to the world frame, and $a$ is the body-frame acceleration. This simplified kinematics model is used solely for policy training. For benchmarking against other methods, the RotorS dynamics model is employed.

The Markov Decision Process (MDP) for our model is defined as a tuple $(\mathcal{S}, \mathcal{A}, \mathcal{P}, \mathcal{R}, \gamma)$, where $\mathcal{S}$ represents the state space, $\mathcal{A}$ denotes the action space, $\mathcal{P}$ defines the transition probability, $\mathcal{R}$ is the reward function, and $\gamma$ signifies the discount factor. The state space $\mathcal{S}$ comprises the current latent representation $\mathbf{z}_t$, along with the drone's state and target information $\mathbf{x}_t$. The action space $\mathcal{A}$ includes the acceleration in the body frame and the yaw rate. The transition probability $\mathcal{P}$ is modeled by the kinematics equation as described in Equation \ref{eq3}.

\begin{figure*}[thpb]
   \centering
   \subcaptionbox{}{%
   {}\vspace{0.5cm}\includegraphics[width=2.2in]{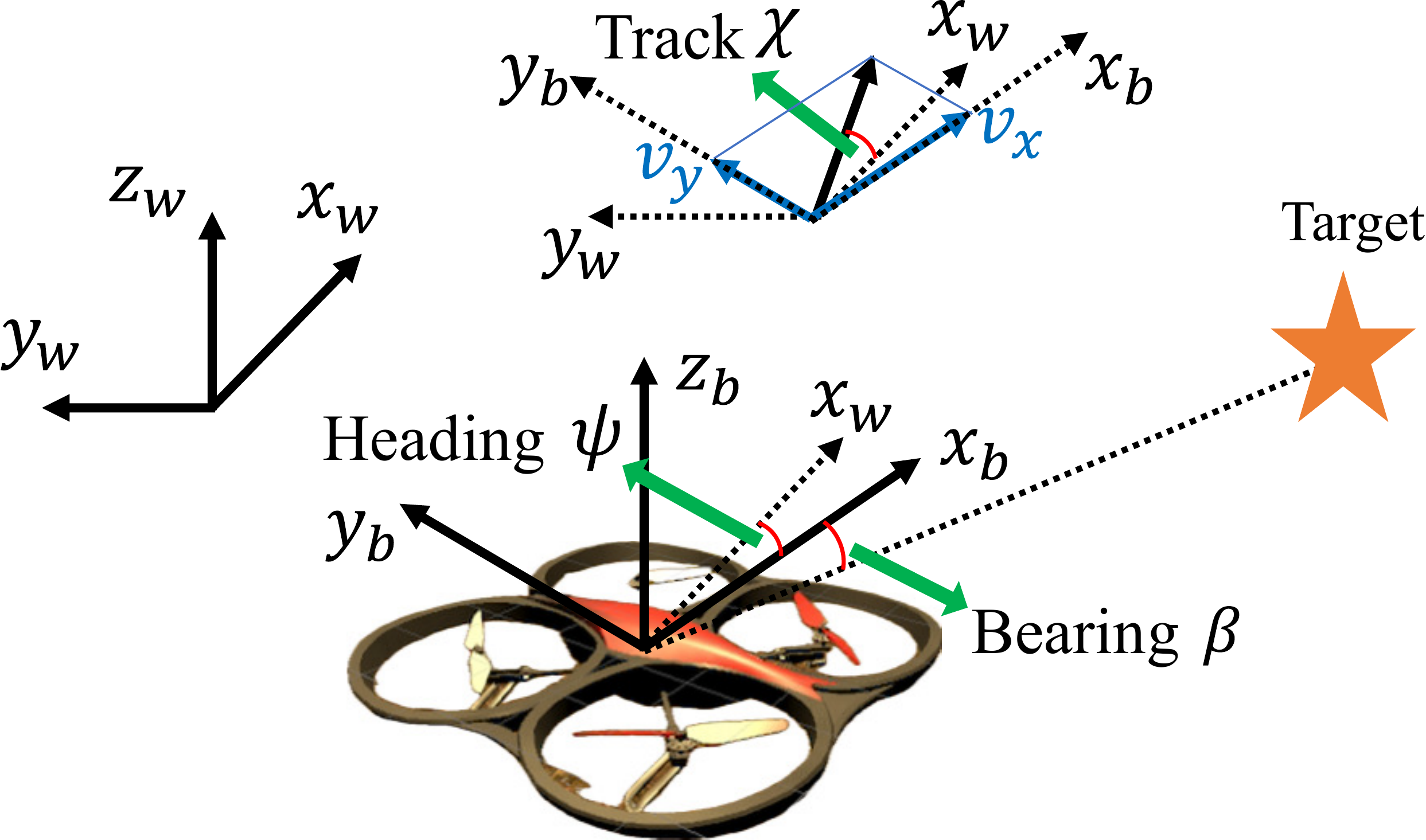}}
   \subcaptionbox{}{%
   \includegraphics[width=2.2in, trim=80 150 0 90, clip]{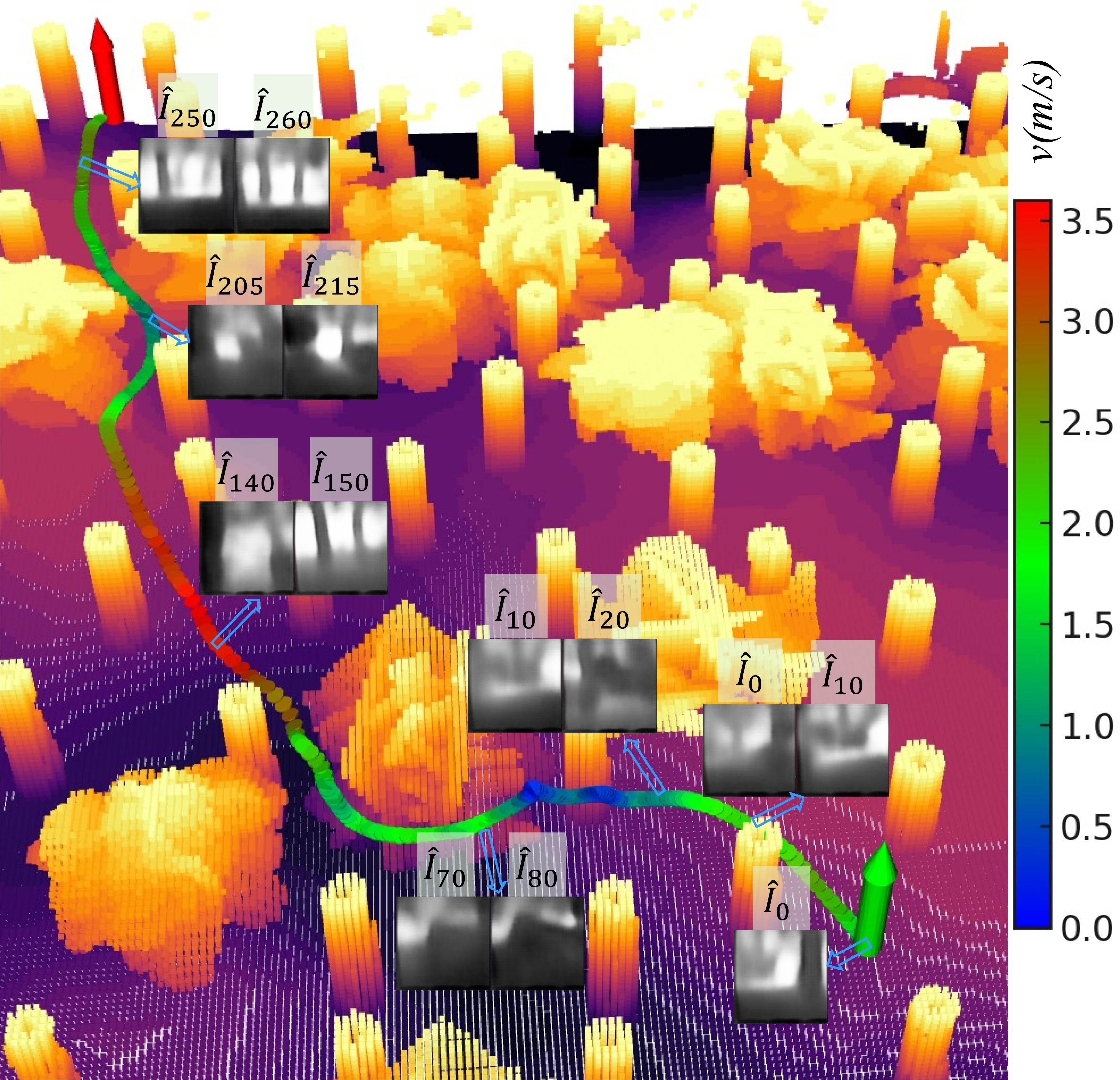}}
   \subcaptionbox{}{%
   \includegraphics[width=2.4in, trim=0 0 30 0, clip]{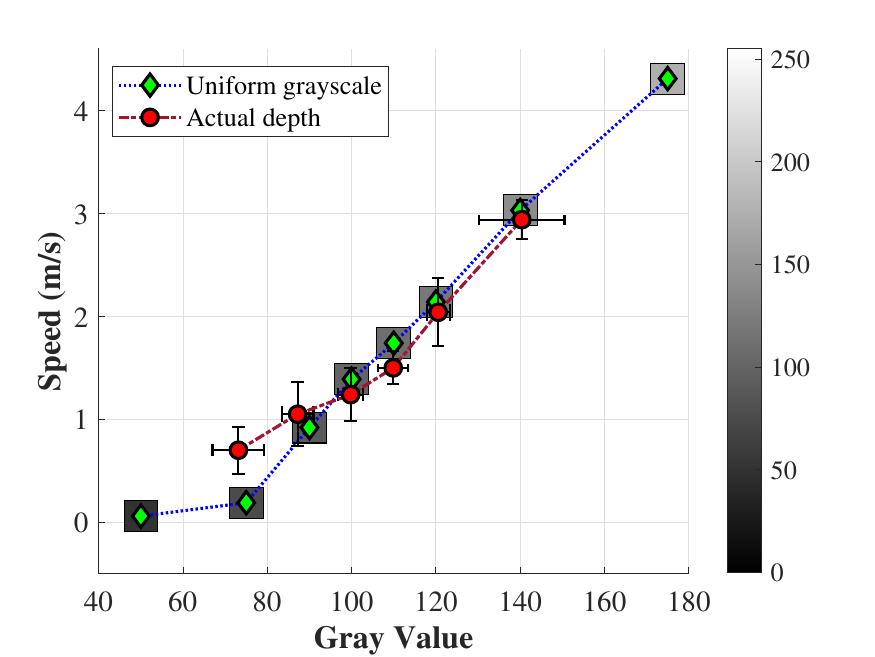}}
   \caption{(a) illustrates the drone's coordinate system, with the bearing angle $\beta$ between body axis $x_b$ and the target vector, and the track angle $\chi$ as the horizontal velocity's direction in the world frame. (b) presents an adaptive drone trajectory in a cluttered environment. The drone decelerates when navigating complex obstacles and accelerates in simpler scenarios, demonstrating dynamic speed adjustment based on obstacle density. (c) displays the drone's average speed in response to uniformly bright gray images (green diamond), and the mean and standard deviation of the speed when the input are actual depth images from (b) (red round).}
   \label{figure4}
\end{figure*}

As illustrated in Figure \ref{figure2}(a), the drone's state and target information, denoted as $\mathbf{x}_t$, is represented by a vector of length $N_s$. In our specific case, $N_s$ is equal to 7. The coordinate system is shown in Figure \ref{figure4}(a): the drone's current position and velocity are denoted as $p(x_t, y_t, z_t)$ and $v(v_x, v_y, v_z)$, respectively, with the target position represented as $p_g(x_g, y_g, z_g)$. The vector from the drone to the target is $\mathbf{d} = p_g - p$, and the drone's heading angle is $\psi$. The bearing angle $\beta$ is defined as the angle between the body frame's $x_b$ axis and the target vector, while the track angle $\chi$ represents the direction of horizontal velocity in the world frame. The drone's state and target information $\mathbf{x}_t$ is defined as:
\begin{equation}
   \begin{aligned}
   \mathbf{x}_t &= [d_{\text{hor}}, v_{\text{hor}}, \beta^{\prime}, d_z, v_z, \chi^{\prime}, \psi], \\
   d_{\text{hor}} &= \ln(\sqrt{d_x^2 + d_y^2} + 1), \quad
   v_{\text{hor}} = \sqrt{v_x^2 + v_y^2}, \\
   \beta^{\prime} &= \beta + \psi = \arctan\left(\frac{d_y}{d_x}\right), \\
   \chi^{\prime} &= \chi - \psi = \arctan\left(\frac{v_y}{v_x}\right),
   \end{aligned}
   \label{eq4}
\end{equation}
   
where $d_{\text{hor}}$ is the logarithm of the horizontal distance from the drone to the target, $v_{\text{hor}}$ is the drone's horizontal velocity,  $v_z$ is the vertical velocity, $\beta^{\prime}$ represents drone-to-target direction in world frame, $\chi^{\prime}$ indicates the velocity direction in body frame, and $d_z$ is the vertical distance to the target.

In our experiments, the horizontal distance to the target is usually much longer than the vertical distance. Therefore, we use different state information and reward functions for horizontal and vertical movements.

\subsection{Reward Functions}

The reward function is designed to ensure that the drone flies safely and efficiently. As reaching the target and avoiding collisions are sparse rewards, we introduce a progressive reward to efficiently guide the drone. The progressive reward is defined as follows:
\begin{equation}
   \begin{aligned}
   r_{\text{progress}} = & \lambda_d \cdot d_{\text{hor}} + \lambda_b \cdot \left|\chi^{\prime} + \psi - \beta^{\prime}\right| + \\
   & \text{sign}( v_{\text{hor}} - v_{\text{max}}) \cdot \lambda_v \cdot v_{\text{hor}} + \lambda_z \cdot d_z + \\
   & \lambda_f \cdot |\chi^{\prime}| + \lambda_a \cdot \|\mathbf{a}_{t-1} - \mathbf{a}_t\|, \\
   \end{aligned}
   \label{eq5}
\end{equation}
where $\lambda_d$, $\lambda_b$, $\lambda_v$, $\lambda_z$, $\lambda_f$, and $\lambda_a$ are weights for each term. $\mathbf{a}_t$ is the acceleration output of policy at time $t$. The first two terms guide the drone towards the target, penalizing horizontal distance and encouraging correct directionality. The third term penalizes horizontal velocity, activated only when $v_{\text{hor}} > v_{\text{max}}$ (with $\lambda_v = 0$ for $v_{\text{hor}} < v_{\text{max}}$). The fourth term penalizes vertical distance, while the fifth term encourages forward flight. The final term, penalizing jerk, ensures smoother flight. 

Then the whole reward function is defined as:
\begin{equation}
   \begin{aligned}
   r = \begin{cases}
   r_\text{exceed} & \text{if } (p_t<p_{min} \text{ or } p_t>p_{max}) \\
   & (p \in \{x,y,z\}) \\
   \dfrac{r_\text{arrive}}{TRAV} & \text{if } \|\mathbf{d}\|<d_{min} \\
   r_\text{collision} & \text{if } \text{collision} \\
   r_\text{progress} & \text{otherwise} \\
   \end{cases} \\
   \end{aligned}
   \label{eq6}
\end{equation}
We define $r_{\text{exceed}}$ as the boundary-exceed penalty, $r_{\text{arrive}}$ as the target arrival reward, and $r_{\text{collision}}$ as the collision penalty. $TRAV$, introduced by Nous et al. \cite{nous2016performance}, measures environmental clutter, accounting for drone's size and complex obstacle shapes. Higher values indicate easier navigation. PPO, trained within a fixed time window, incentivizes faster flight for higher arrival rewards. To balance safety and agility, the arrival reward inversely correlates with $TRAV$, while the horizontal velocity penalty in Equation \ref{eq5}, moderates speed.

The episode terminates once any of the previously mentioned conditions are met, following which the drone is reset to a new random starting point. In our setup, the values are configured as follows: $r_\text{exceed}=-2.0$ for exceeding boundaries, $r_\text{arrive}=10.0$ for reaching the target, and $r_\text{collision}=-2.0$ for collisions. The traversability range, $TRAV$, is set between 3 and 13. The progressive reward, $r_\text{progress}$, ranges from -0.2 to 0 as defined in Equation \ref{eq5}. Notably, this progressive reward is considerably smaller than the other rewards.

\subsection{Training in Varying Complexity Environments}
We utilize AvoidBench \cite{10161097} to construct the RL environment. AvoidBench is tailored for evaluating vision-based obstacle avoidance algorithms, offering customizable environmental complexity through the radius of the Poisson distribution. AvoidBench is based on Flightmare \cite{song2020flightmare}, but additionally allows for the inclusion of large bushes as obstacles, which present more variability than thin red trees typically found in Flightmare's scenes.

To increase training efficiency, we begin with a warm-up period for the policy network in a simpler environment (12-meter Poisson distribution radius), facilitating easy target navigation learning and high arrival rewards. We then raise the complexity (Poisson radius [3.0, 5.4] meters), effectively teaching the drone to adjust its speed based on environmental complexity—accelerating in simpler areas and slowing down in denser ones.

As illustrated in Figure \ref{figure4}(b), the task involves the drone flying from the green arrow (start point) to the red arrow (target). The trajectory is color-coded to represent the drone's velocity.
Additionally, we display some predicted depth images generated by our memory-augmented latent representation. These images are presented as pairs of $(\hat{I}_{t-10}, \hat{I}_{t})$. For instance, in the pairs $(\hat{I}_{0}, \hat{I}_{10})$ and $(\hat{I}_{10}, \hat{I}_{20})$, it is evident that the drone retains memory of the depth image $\hat{I}_{10}$ seen 10 timestamps earlier. Observations from pairs $(\hat{I}_{10}, \hat{I}_{20})$, $(\hat{I}_{70}, \hat{I}_{80})$, and $(\hat{I}_{205}, \hat{I}_{215})$ demonstrate the drone's tendency to decelerate when encountering complex obstacles and to accelerate when observed distances in the flight direction are larger, as seen in $(\hat{I}_{140}, \hat{I}_{150})$ and $(\hat{I}_{250}, \hat{I}_{260})$. 

For the trajectory shown in Figure \ref{figure4}(b), depth maps captured every 0.1 seconds are categorized into six groups based on their average gray values. The mean and standard deviation of speed for each depth map group are depicted in Figure \ref{figure4}(c) (red circles). Additionally, the response speed to uniformly bright gray images is displayed as green diamonds, calculated during periods of stable acceleration at zero. This closely matches the speed when navigating actual depth. The plots illustrate an inverse relationship between drone speed and obstacle density, indicating faster speeds in less cluttered environments and slower speeds amidst denser obstacles.

\section{Experiments}

To verify the effectiveness of MAVRL, we conducted a series of experiments. We trained the different results of predicting past, current and future depth respectively for the memory-augmented latent representation, and then compared the performance of different latent representations by training the policy network with the same parameters. To verify whether the varying speed ability of MAVRL can improve the success rate and get a good balance between safety and agility, we also trained the policy network with a fixed speed, comparing both of them with a state-of-the-art method: Agile-Autonomy \cite{Loquercio2021Science}. Finally, we deployed our network on a real drone with minimal fine-tuning.

All simulation experiments are run on a server with an Intel Core i7-13700K CPU and an NVIDIA GeForce RTX 4090 GPU. To get enough training results for statistical analysis, we create 5 docker containers in the server and run per configuration 5 parallel, independent training processes.
\subsection{Latent Representation}

To assess the efficacy of the memory-augmented latent representation, we conducted an experiment where the LSTM was trained with various reconstruction configurations. These different configurations were then utilized to train the policy network. We examined seven distinct types of latent representations for this purpose:
\begin{itemize}
   \item Current depth image prediction, $I_t$,
   \item Future depth prediction, $I_{t+10}$,
   \item Current depth, $I_t$, with short-term future, $I_{t+10}$,
   \item Current depth, $I_t$, with long-term future, $I_{t+20}$,
   \item Current depth, $I_t$, and past depth, $I_{t-10}$,
   \item Current depth, $I_t$, with more distant past, $I_{t-20}$,
   \item Current depth, $I_t$, combined with past, $I_{t-10}$, and future, $I_{t+10}$,
\end{itemize}
where predicting current depth \cite{kulkarni2023semantically} and predicting future depth \cite{9385894} separately are the most common 
in the literature.

\begin{figure}[thpb]
   \centering
   \includegraphics[width=3.2in, trim=20 20 40 20, clip]{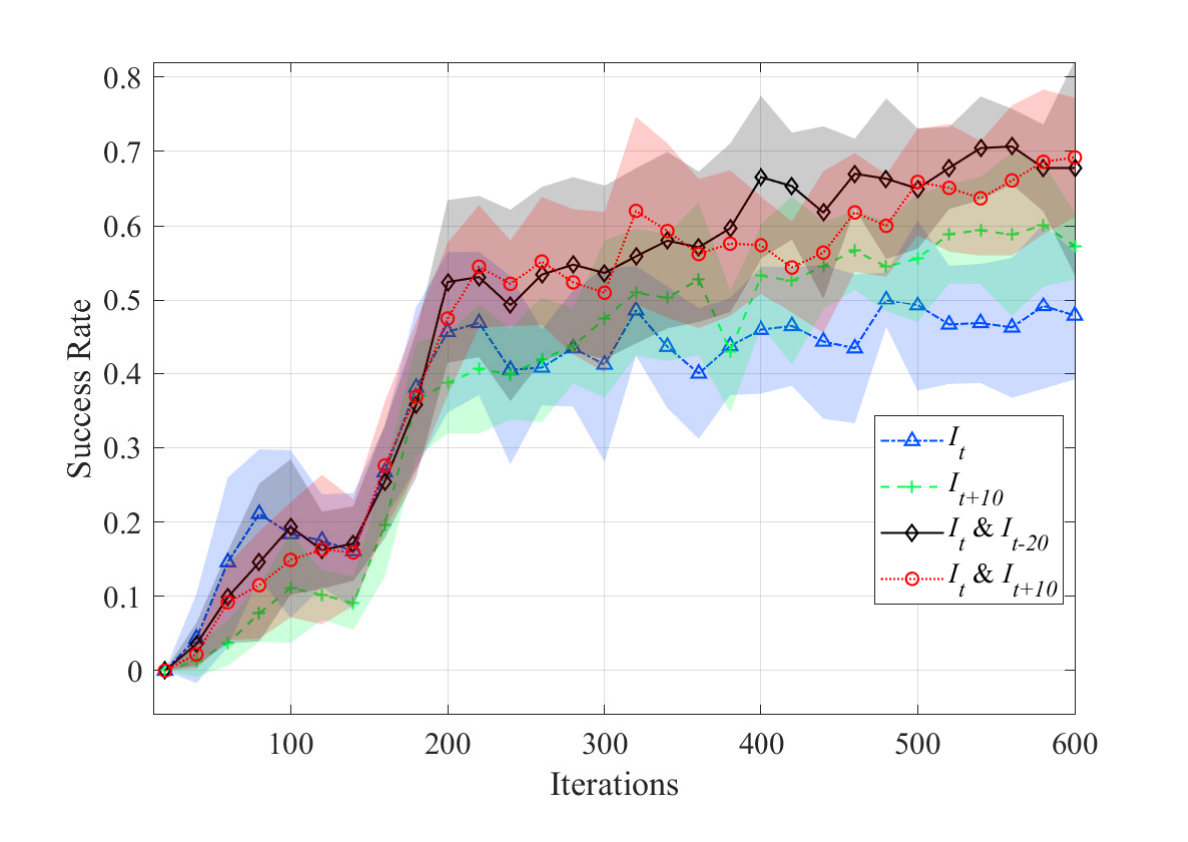}
 \caption{Success rates of $I_t$, $I_{t+10}$, and the superior combinations $I_t \& I_{t-20}$ and $I_t \& I_{t+10}$. The shadow area represents the standard deviation.}
   \label{figure6}
\end{figure}

To thoroughly evaluate how various latent representations affect policy network performance and the benefits of augmented memory, we conducted an extensive testing regimen. The policy network, using consistent latent representations, was trained ten times, each with a unique random seed, saving checkpoints every 20 iterations for a total of 600 iterations. We then utilized these checkpoints to test the policy network on 4 different maps, conducting 25 trials per map, amounting to $30 \times 4 \times 25$ trials per latent space. The results of these trials are detailed in Figure \ref{figure6} and Table \ref{table1}.

Figure \ref{figure6} displays the success rates of $I_t$, $I_{t+10}$, and the superior combinations $I_t \& I_{t-20}$ (the better one compared with $I_t \& I_{t-10}$) and $I_t \& I_{t+10}$ (the better one compared with $I_t \& I_{t+20}$). We compared the highest success rate checkpoints for each latent representation, detailing average success rates and standard deviations in Table \ref{table1}. The Permutation Test \cite{boik1987fisher} with 10,000 permutations was used to validate our memory-augmented latent representation against commonly used $I_t$ and $I_{t+10}$ methods. A P-value below 0.05 indicates significant differences between latent representations. Thus, we deduce that augmenting current depth with past or future information consistently outperforms predictions based solely on current depth. The combination of $I_t$ and $I_{t-20}$ emerged as the most effective, closely followed by the combination of current and future depth $I_t \& I_{t+10}$ which are both much better than only predicting future depth. Predicting past $I_{t-10}$, current $I_t$, and future $I_{t+10}$ also performed well but didn't notably surpass the $I_t$ and $I_{t-20}$ combination.

\begin{table}[thpb]
   \caption{Comparison of different types of latent space.}
   \vspace{-2mm}
   \label{table1}
   \begin{center}
   \begin{tabular}{cccc}
   \toprule
   \multirow{2}{*}{Latent space} & \multirow{2}{*}{$\text{Mean}\pm\text{std}$} & \multicolumn{2}{c}{P-Value} \\
   \cline{3-4}
   & & (for $I_t$) & (for $I_{t+10}$) \\
   \midrule
   $I_t$ & $0.500 \pm 0.037$ & - & -\\ 
   $I_{t+10}$ & $0.601 \pm 0.083$ & 0.0015 & - \\
   $I_t \& I_{t-10}$ & $0.636 \pm 0.089$ & 0.0001 & 0.1913 \\
   $I_t \& I_{t-20}$ & $\textbf{0.707} \pm \textbf{0.051}$ & 0.0 & 0.0016 \\
   $I_t \& I_{t+10}$ & $0.692 \pm 0.080$ & 0.0 & 0.0110 \\
   $I_t \& I_{t+20}$ & $0.664 \pm 0.100$ & 0.0001 & 0.0680 \\
   $I_{t-10} \& I_t \& I_{t+10}$ & $0.649 \pm 0.102$ & 0.0005 & 0.1239 \\
   \bottomrule
   \end{tabular}
   \end{center}
   \vspace{-6mm}
\end{table}

Given that the LSTM training dataset was gathered using an initial policy, we compiled Table \ref{table2} to illustrate the variations in predicting past, current, and future depth across different latent spaces before or after retraining with PPO. The mean and standard deviation are derived from the gray scale values of the depth images. It can be seen from the results that the error in predicting the future is the most obvious. And all reconstruction effects decrease to a small extent after retraining PPO.

\begin{table}[thpb]
   \caption{Errors of different prediction items.}
   \vspace{-2mm}
   \label{table2}
   \begin{center}
   \begin{tabular}{ccc}
   \toprule
   \multirow{2}{*}{Predicted Item} & \multicolumn{2}{c}{$\text{Mean}\pm\text{std}$} \\
   \cline{2-3}
   & (Before PPO retraining) & (After PPO retraining) \\
   \midrule
   $I_{t-20}$ & $30.70 \pm 13.11$ & $32.93 \pm 17.28$ \\
   $I_{t-10}$ & $24.51 \pm 10.82$ & $26.49 \pm 14.78$ \\
   $I_{t}$ & $18.73 \pm 8.39$ & $18.97 \pm 10.15$ \\
   $I_{t+10}$ & $51.03 \pm 27.27$ & $53.79 \pm 30.18$ \\
   $I_{t+20}$ & $58.51 \pm 28.58$ & $58.70 \pm 32.89$ \\
   \bottomrule
   \end{tabular}
   \end{center}
   \vspace{-3mm}
\end{table}

For memory-augmented latent spaces, we found they excel in environments with large obstacles. Figure \ref{figure1}(b) illustrates a clear performance disparity between latents with and without memory. Latents lacking memory often led to drones becoming entrapped in large obstacles. Conversely, drones with memory-augmented latents adeptly circumnavigated these obstacles, choosing longer yet safer paths to their targets. This underscores memory's vital role in improving drones' obstacle avoidance, especially amidst larger obstructions.

\subsection{Benchmarking for Varying Speed Policy}

\begin{figure*}[thpb]
   \centering
     \includegraphics[width=6.8in, trim=10 0 100 0, clip]{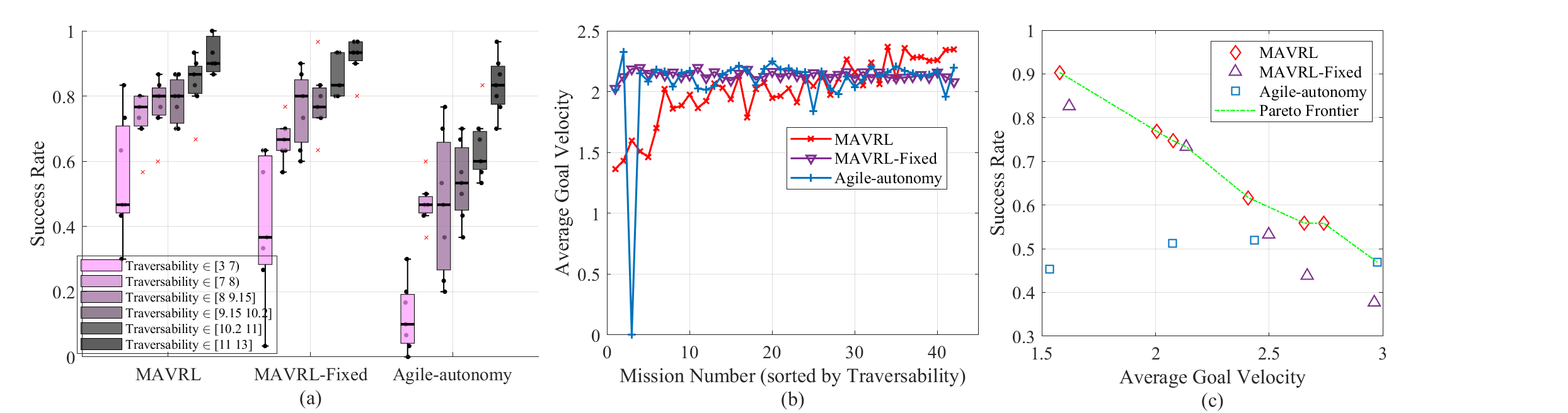}
 \caption{(a) is the success rate of 2 different MAVRL versions and Agile-Autonomy. (b) is the average goal velocity (AGV) of 2 different MAVRL versions and Agile-Autonomy. (c) is the Pareto frontier of success rate versus average flight speed.}
   \label{figure7}
\end{figure*}

To assess the impact of a varying speed policy, we conducted an experiment where a policy network trained with variable speeds was compared against one operating at a fixed speed. For the fixed speed setup, we adjusted the velocity penalty in the reward formula (Equation \ref{eq5}) to $\lambda_v \cdot |v_{\text{hor}} - v_{\text{desire}}|$, with $v_{\text{desire}}$ representing the desired velocity. Although $\lambda_v$ was set high, it remained lower than the collision penalty, allowing consistent speed learning. This fixed speed model was then benchmarked against the advanced Agile-Autonomy method \cite{Loquercio2021Science}, utilizing the AvoidBench framework \cite{10161097}.

For our experiment, MAVRL utilized the MPC controller from Agilicious \cite{foehn2022agilicious} when outputting acceleration commands, which then generated body rate and thrust commands for the drone. Conversely, Agile-Autonomy generated a set of waypoints, with a polynomial curve fitting these points, and the high level controller is the same MPC.

The benchmarking outcomes in Figure \ref{figure7} compare MAVRL with varying and fixed speeds, and Agile-Autonomy, all using a memory-augmented latent representation. Figure \ref{figure7}(a) shows the success rates over six groups, each with seven maps and 30 trials per map, indicating MAVRL with varying speed often performs best. Figure \ref{figure7}(b) explores the link between average goal velocity (AGV) \cite{10161097} and map complexity, revealing MAVRL with varying speed tends to have higher AGV in less complex environments, while all algorithms maintain a similar AGV (around 2.0 m/s).

\begin{figure}[thpb]
   \centering
   { {}\vspace{-0.5mm}
   \includegraphics[width=3.0in]{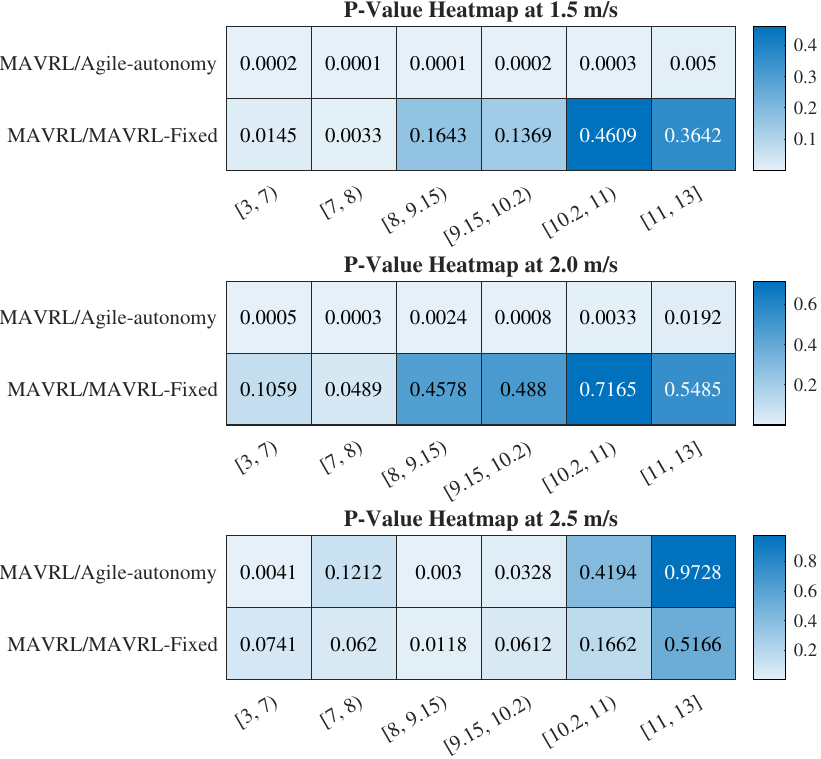}}
   \caption{Permutation Test Results. The p-values indicate the statistical significance of the success rate distributions of the different MAVRL versions and Agile-Autonomy.}
   \label{figure8}
   \vspace{-3mm}
\end{figure}

To validate MAVRL's superior performance across various agility levels when employing varying speeds, we fine-tuned the reward function parameters of both MAVRL variants to achieve different average flight speeds. This led to the construction of a Pareto frontier of success rate versus average flight speed, as shown in Figure \ref{figure7}(c). The results confirm that MAVRL with varying speed forms the Pareto frontier, dominating the results of the other methods. However, its average speed could not exceed 3.0 m/s due to flight distance limitations, although the maximum speed reached 5.5 m/s.

While MAVRL with varying speed showed a higher average success rate than the fixed speed variant in Figure \ref{figure7}(a), the difference was not substantial across all scenarios, particularly in specific complexity environments (e.g., traversability between 8 and 9.15 or 9.15 and 10.2). To further validate these findings, we also conducted a Permutation Test to compare the success rate distributions of the different MAVRL versions and Agile-Autonomy. The results are presented in Figure \ref{figure8} with values below 0.05 indicating statistical significance. Our analysis revealed that MAVRL with varying speed significantly outperforms the fixed speed version in cluttered environments (traversability between 3.0 and 8.0) and also shows a notable difference compared to Agile-Autonomy in most environments.

\subsection{Real World Tests}

\begin{figure}[thpb]
   \centering
   {\includegraphics[width=3.2in,trim={0 0.6cm 0 5cm},clip]{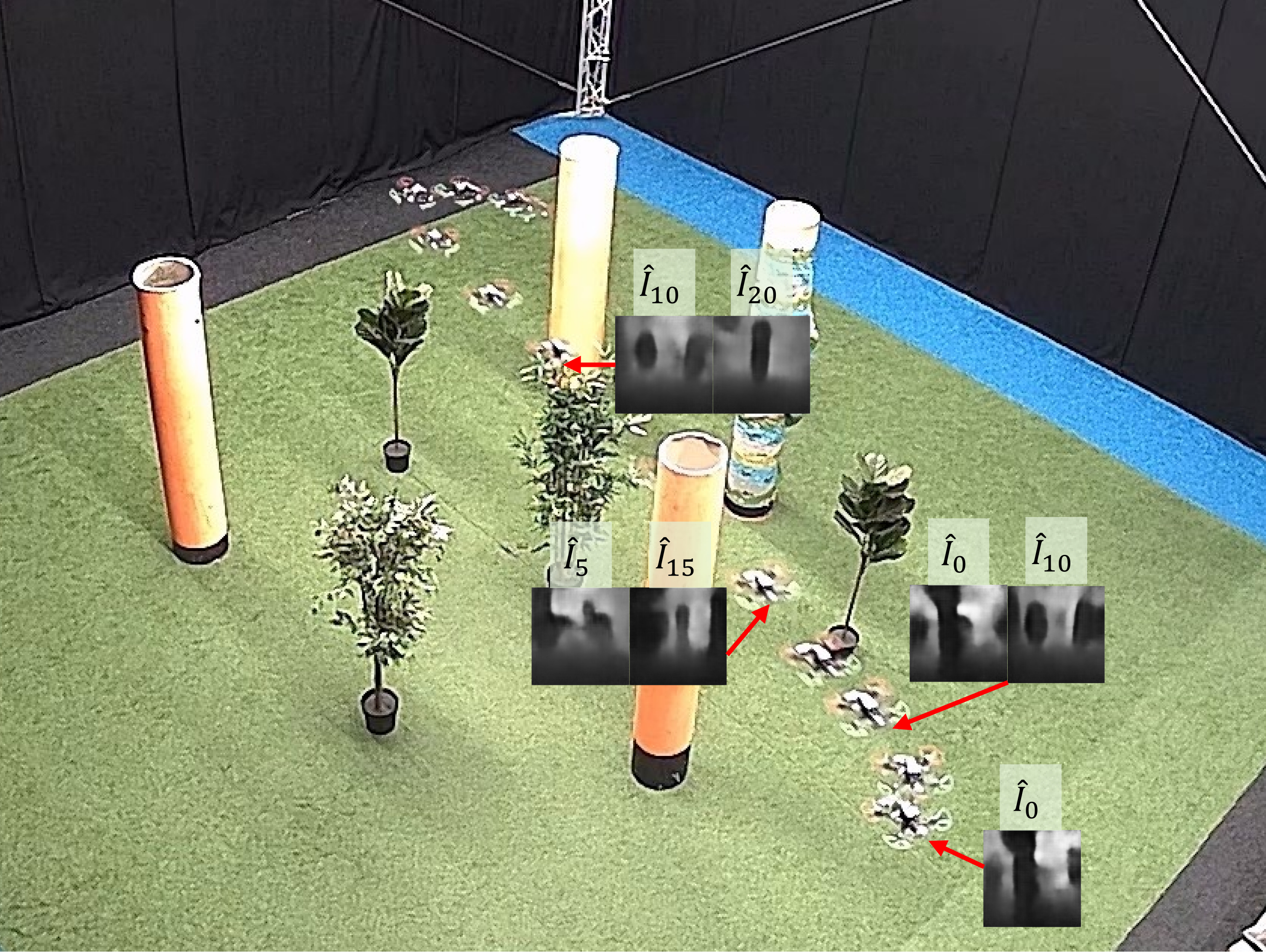}}
   \caption{Real world test of MAVRL. Then latent representation is shown as a pair of $(\hat I_{t-10}, \hat I_{t})$.}
   \label{figure9}
   \vspace{-3mm}
\end{figure}

To validate MAVRL's real-world efficacy, we implemented our network on an actual drone, maintaining the same architecture and hyperparameters as in our simulation experiments. Due to the real scene's environmental background shown in Figure \ref{figure9} being too close to the obstacles, the VAE and LSTM trained in simulation struggled to differentiate between obstacles and background effectively. Consequently, the VAE and LSTM components were fine-tuned with real-world data while maintaining the original policy parameters. Our test setup included a quadrotor equipped with 5-inch propellers and a Realsense D435i camera, powered by a Jetson Xavier NX featuring a 384-core GPU, 48 Tensor Cores, and a 6-core ARM CPU. Depth images were generated using the Realsense D435i stereo camera, while an Optitrack motion capture system provided accurate ground truth data for the drone's state. As shown in Figure \ref{figure9}, the drone successfully navigated a cluttered environment, utilizing a latent representation augmented with past memory. Additional test results are available in our supplementary video. This real-world performance mirrors our simulation results, demonstrating the drone's capability to navigate safely and efficiently.

\section{CONCLUSION}

Our approach leverages memory-augmented latent representations to endow the drone with a recollection of past scenarios. Experimental results demonstrated that reconstructing a more extensive history of past and current depth information significantly enhances the drone's performance in reinforcement learning-based obstacle avoidance tasks. Additionally, we established that adopting a varying speed strategy not only improves success rates but also strikes an optimal balance between safety and agility. The successful deployment of our network on a real drone, requiring minimal fine-tuning, marks a significant achievement. Looking forward, our aim is to delve into the development of a fully end-to-end methodology that enables the policy network to directly generate low-level control commands.






\bibliographystyle{IEEEtran}
\bibliography{IEEEexample}

\end{document}